\documentclass{article}

\PassOptionsToPackage{numbers, compress}{natbib}
\usepackage[dblblindworkshop, final]{neurips_2026}\workshoptitle{AXIOM 2026: Foundations of Efficient Deep Learning --- Towards predictive principles for efficient AI connecting deep learning theory, scaling laws, and efficiency}

\usepackage[utf8]{inputenc}
\usepackage[T1]{fontenc}
\usepackage{hyperref}
\usepackage{diagbox}
\usepackage{url}
\usepackage{placeins}
\usepackage{algorithm}
\usepackage{algpseudocode}
\usepackage{booktabs}
\usepackage{booktabs}
\usepackage{amsmath, amssymb}
\usepackage{amsfonts}
\usepackage{nicefrac}
\usepackage{microtype}
\usepackage{xcolor}
\usepackage{makecell}
\usepackage{multirow}
\usepackage{graphicx}

\title{TopK-Guided: Adaptive, Budget-Aware Activation Sparsity for Efficient LLM Inference}

\author{%
  Mukund Agarwalla \\
  MBZUAI \\
  \texttt{mukund.agarwalla@mbzuai.ac.ae}
  \And
  Chih-Jen Lin \\
  MBZUAI \\
  National Taiwan University \\
  \texttt{ChihJen.Lin@mbzuai.ac.ae}
}

\begin{document}

\maketitle

\begin{abstract}
Activation sparsity speeds up large language model (LLM) inference by setting unimportant activations to zero so that the corresponding computations can be skipped. Existing training-free methods, however, make different trade-offs: threshold-based methods such as TEAL adapt the sparsity level to each token but do not tightly control the realised sparsity, while TopK-based methods such as WINA enforce a fixed sparsity level but use the same sparsity budget for every token. Both also apply the same budget across transformer blocks, despite large differences in block sensitivity. We introduce \textbf{TopK-Guided}, a training-free method that addresses both limitations by combining bounded token-level sparsity adaptation with sensitivity-aware block-level budget allocation. Across Llama-2 and Llama-3 models, TopK-Guided consistently improves perplexity and downstream accuracy over TEAL and WINA while preserving essentially the same sparsity-dependent projection compute as WINA, with the largest gains at high sparsity. Ablations show that both components provide complementary improvements.
\end{abstract}

\section{Introduction}
\label{sec:intro}

Every token generated by a large language model (LLM) requires a full forward pass through the network, and within each transformer block most of this cost comes from a few large linear projections ($W_q, W_k, W_v, W_o$ in attention and $W_{\text{gate}}, W_{\text{up}}, W_{\text{down}}$ in the MLP), each multiplying a weight matrix by an activation vector \cite{liu2023dejavucontextualsparsity}. A well-documented property of trained LLMs is that many of the activation entries feeding these projections are already close to zero at inference time \cite{li2023lazyneuronphenomenonemergence} and contribute negligibly to the output \cite{liu2023dejavucontextualsparsity, mirzadeh2023relustrikesbackexploiting}.
 
Activation sparsity exploits this. Before a projection $Wh$, it sets the smallest-magnitude entries of the activation vector $h$ to zero; whenever an entry $h_j$ is zeroed, the corresponding column $W_{:,j}$ can be skipped, since it only ever multiplies $h_j$. Zeroing a fraction $s$ of the entries thus skips a fraction $s$ of the projection's work. Unlike weight pruning \cite{sun2024simpleeffectivepruningapproach, frantar2023sparsegptmassivelanguagemodels}, which permanently removes the same parameters for every input, activation sparsity leaves the model untouched and chooses its sparsity pattern afresh for each token \cite{liu2023dejavucontextualsparsity}.
 
Given a target sparsity $s$, a method must decide, for each token, which entries of $h$ to zero. Two recent methods define the state of the art and adopt opposite strategies. TEAL \cite{liu2025trainingfreeactivationsparsitylarge} calibrates, per projection, a single threshold $\tau$ offline so that on average a fraction $s$ of activation magnitudes fall below it, and at inference retains every entry above the threshold:
\begin{equation}
\tilde{h}_j =
\begin{cases}
h_j, & |h_j| \ge \tau,\\[2pt]
0,   & |h_j| < \tau.
\end{cases}
\end{equation}
At inference time, the same threshold $\tau$ affects each token differently: a token whose activations are mostly small loses many entries, giving it higher sparsity, while a token with larger activations keeps more, giving it a lower sparsity. The mask therefore adjusts itself to each token, a property we call \emph{token adaptivity}. This adaptivity comes at a cost: since $\tau$ is fixed after calibration but activation distributions vary across tokens, the realised sparsity drifts from the target $s$ (Appendix~\ref{app:threshold-vs-topk}), and the resulting speedup becomes unpredictable. WINA \cite{chen2026winaweightinformedneuron} takes the opposite strategy: rather than fixing a threshold and letting the number of kept entries vary, it fixes that number directly through TopK selection. For each token it keeps the $k = \lfloor d(1-s) \rfloor$ entries with the largest weight-informed score $|h_j| \cdot \lVert W_{:,j} \rVert_2$, which ranks entries by how much each one moves the output, and zeroes the rest ($d$ is the number of entries in $h$). Since $k$ is the same for every token, exactly a fraction $s$ is zeroed each time, so the realised sparsity always matches the target. But fixing $k$ removes the token adaptivity that TEAL had for free. The two methods therefore sit at opposite ends of a single trade-off. TEAL adapts to each token but cannot control the realised sparsity, while WINA controls the sparsity exactly but cannot adapt to the token. Ideally we would want both: a mask that adjusts to each token while keeping the realised sparsity close to the target.
 
\begin{figure}[t]
    \centering
    \includegraphics[width=0.7\linewidth]{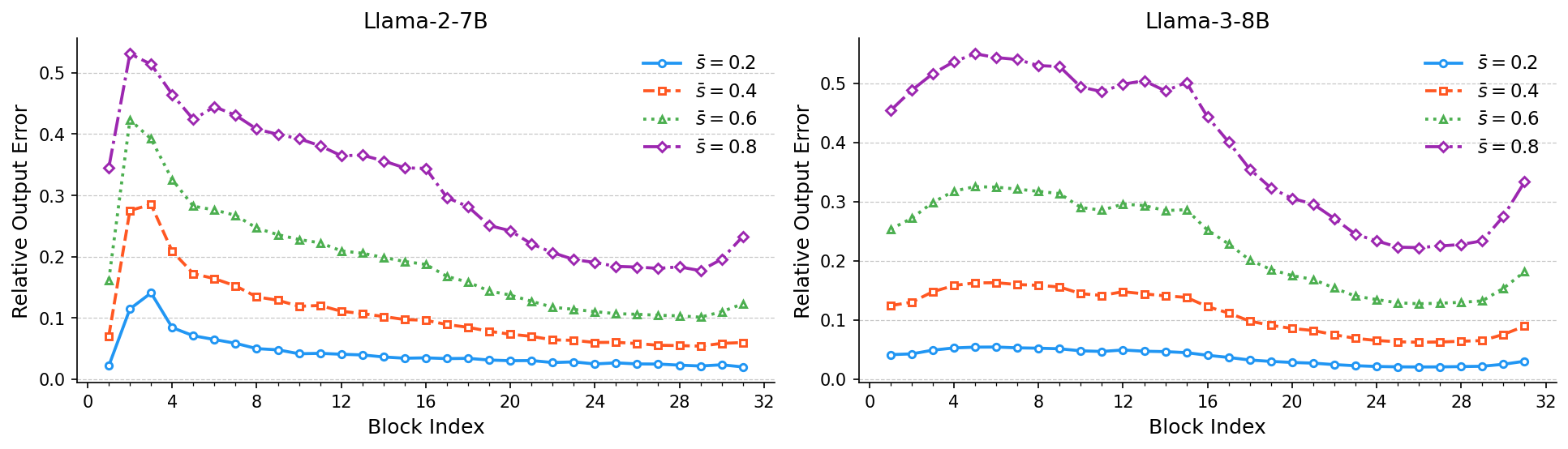}
    \caption{Per-block sensitivity $e_i$ (Eq.~\ref{eq:sensitivity}) on Llama-2-7B and Llama-3-8B, shown at four probe levels $\bar{s} \in \{0.2, 0.4, 0.6, 0.8\}$. }
    \label{fig:sensitivity}
\end{figure}

Both TEAL and WINA share a further limitation: they assign the same budget $s$ to every transformer block. Blocks, however, are not equally tolerant of sparsification: sparsifying each block on its own and measuring how much its output changes shows that early blocks are far more sensitive than later ones (Figure~\ref{fig:sensitivity}). A uniform budget therefore over-sparsifies the fragile early blocks and under-uses the robust later ones.
 
We propose \textbf{TopK-Guided}, two complementary training-free modifications to WINA, one for each issue: \textbf{1) TopK-Guided masking} (Section~\ref{sec:topk-guided}) recovers the per-token adaptivity that WINA gives up while keeping the realised sparsity within a bounded $\pm\delta$ of the target. We reuse TEAL's calibrated threshold $\tau$ to estimate how sparse each token naturally is, then clip that estimate to a narrow band $\pm\delta$ around $s$, before applying TopK. \textbf{2) Sensitivity-aware block sparsity} (Section~\ref{sec:block}) replaces the uniform budget with one matched to each block's sensitivity: a single calibration pass measures it, and sensitive blocks then get less, robust ones more. Like OWL\citep{yin2025outlierweighedlayerwisesparsity}, we allocate sparsity non-uniformly across blocks, but base it on a direct sensitivity measurement rather than a weight-outlier signal.

Across Llama-2-7B \cite{touvron2023llama2openfoundation} and Llama-3-8B \cite{grattafiori2024llama3herdmodels}, TopK-Guided beats both TEAL and WINA on perplexity and downstream tasks at matched compute, with the largest gains at high sparsity.

\section{Method}
\label{sec:method}

We build our method on top of WINA, keeping its tensor rotation of the
residual stream (Appendix~\ref{app:rotation}) and its greedy
per-projection allocation within each block (Appendix~\ref{app:greedy})
unchanged; with sparsification applied only during the decode phase. Our
two contributions modify WINA in exactly two places: the masking rule and the block-level budget assignment. Table~\ref{tab:notation} in Appendix~\ref{app:notation} collects the sparsity-related symbols used throughout, for reference.

\subsection{TopK-Guided Masking}
\label{sec:topk-guided}

Threshold masking adapts to each token but its realised sparsity can
deviate from the target in an unbounded, input-dependent way, making
inference-time speedup unpredictable. TopK masking eliminates this
deviation by enforcing exactly $k$ active elements per token, but at
the cost of complete rigidity: the same fixed fraction $s$ is applied
to every token regardless of its activation distribution, with no
adaptation whatsoever. We propose TopK-Guided masking, which keeps threshold masking's per-token adaptivity while bounding the sparsity deviation to a small band 
$\pm\delta$ around the target.

The key observation is that the calibration threshold $\tau$ already
encodes the typical activation distribution of each projection. At inference we therefore use $\tau$ to cheaply estimate, in $O(d)$, the token's natural sparsity ($\hat{s}^{(t)}$), the fraction of its entries that fall below $\tau$:
\begin{equation}
    \hat{s}^{(t)} = \frac{1}{d}
    \sum_{j=1}^{d}
    \mathbf{1}\!\left[\,|h_j^{(t)}| < \tau\,\right],
    \label{eq:token-estimate}
\end{equation}
where $h_j^{(t)}$ is the $j$-th activation entry for the current
decode token $t$. This estimate tells us how many entries \emph{want}
to be zero for this specific token. We do not apply $\hat{s}^{(t)}$ directly but clip it to a $\pm\delta$ band around the target, giving the per-token target sparsity ($s^{*(t)}$) that we actually enforce:
\begin{equation}
    s^{*(t)} = \min\!\bigl(s + \delta,\; \max\!\bigl(s - \delta,\;
    \hat{s}^{(t)}\bigr)\bigr).
    \label{eq:clip}
\end{equation}
That is, $s^{*(t)}$ equals $\hat{s}^{(t)}$ when the natural estimate lies
within $[s - \delta, s + \delta]$, and is otherwise pulled to the nearer
boundary. Finally, TopK is applied with the resulting count
$k^{(t)} = \max\bigl(1, \lfloor d(1 - s^{*(t)}) \rfloor\bigr)$ active
elements retained per token. The mask thus adapts to each token's
activation pattern like threshold masking, while the realised sparsity is
guaranteed to stay within $\pm\delta$ of the target; WINA's exact TopK is the special case $\delta =0$. The whole step is $O(d)$ per token: an $O(d)$ comparison to form $\hat{s}^{(t)}$, a scalar clip $s^{*(t)}$, and the selection itself. Since only the top-$k^{(t)}$ mask is needed and not a full ordering, we skip sorting and find the cutoff with quickselect \citep{10.1145/366622.366647} in $O(d)$ , rather than the $O(d\log k)$ partial sort WINA's implementation uses. We fix $\delta = 0.01$ throughout, so the realised per-token sparsity deviates by at most one percentage point from the target: small enough to leave the compute savings essentially unchanged, yet enough room for meaningful per-token adaptation. We therefore treat $\delta$ as fixed rather than a tunable hyperparameter.

\subsection{Sensitivity-Aware Block Sparsity}
\label{sec:block}

We give block $i$ its own budget $s_i$, the target sparsity applied to that block, in two steps: measure how sensitive it is to sparsification, then redistribute the global budget $s$ from sensitive blocks to robust ones.

\textbf{Measuring sensitivity:} For block $i$, we run one calibration pass on a small corpus $D$ with block $i$ sparsified at a fixed probe level $\bar{s}$, and record how much its output moves:
\begin{equation}
\label{eq:sensitivity}
e_i = \frac{1}{|D|} \sum_{x \in D} \frac{\lVert f_i(x) - \hat{f}_i(x;\bar{s}) \rVert_2}{\lVert f_i(x) \rVert_2 + \varepsilon},
\end{equation}
where $f_i(x)$ is the block's dense output, $\hat{f}_i(x;\bar{s})$ its output at probe sparsity $\bar{s}$, and $\varepsilon$ avoids division by zero. A large $e_i$ marks a sensitive block, a small one a robust block (Figure~\ref{fig:sensitivity} plots these $e_i$). The probe $\bar{s}$ is only a measuring stick, not the sparsity any block is finally run at.

\textbf{Redistributing the budget:} Let $\bar{e}$ be the mean sensitivity over the $L$ blocks. Each block's budget is
\begin{equation}
\label{eq:redistribution}
s_i = \min\!\big(0.95,\ \max(0.05,\ s + \lambda(\bar{e} - e_i))\big),
\end{equation}
so a block above average sensitivity ($e_i > \bar{e}$) gets less than the target ($s_i < s$) and a robust block ($e_i < \bar{e}$) takes up the slack, leaving the average at $s$. Here $\lambda \ge 0$ sets how hard to redistribute, and the clamp to $[0.05, 0.95]$ keeps any block from going nearly dense or nearly empty.

Each $s_i$ then feeds WINA's existing greedy per-projection allocation (Appendix~\ref{app:greedy}), precomputed at calibration; we add no new machinery, only change the per-block budget it receives. TopK-Guided introduces three hyperparameters absent from WINA: the clip
margin $\delta$ (Section~\ref{sec:topk-guided}), the probe sparsity
$\bar{s}$, and the redistribution strength $\lambda$. We fix
$\delta = 0.01$ throughout, leaving only $\bar{s}$ and $\lambda$ free.

\section{Experiments}
\label{sec:experiments}

\begin{table}[t]
\centering
\caption{C4 perplexity and average accuracy over
eight tasks (Acc \%) on Llama-2-7B and Llama-3-8B.
 Best per column in bold.}
\label{tab:main}
\small
\begin{tabular}{l l cc cc cc}
\toprule
& & \multicolumn{2}{c}{$s=0.3$}
  & \multicolumn{2}{c}{$s=0.5$}
  & \multicolumn{2}{c}{$s=0.7$} \\
\cmidrule(lr){3-4}\cmidrule(lr){5-6}\cmidrule(lr){7-8}
Model & Method & PPL & Acc & PPL & Acc & PPL & Acc \\
\midrule
\multirow{3}{*}{Llama-2-7B}
 & TEAL          & 6.92 & 64.46 & 7.57 & 61.88 & 15.85 & 50.19 \\
 & WINA          & 6.90 & 64.63 & 7.34 & 63.33 & 10.72 & 56.54 \\
 & TopK-Guided   & \textbf{6.89} & \textbf{64.71} & \textbf{7.32} & \textbf{63.42} & \textbf{10.15} & \textbf{56.80} \\
\midrule
\multirow{3}{*}{Llama-3-8B}
 & TEAL          & 9.05 & 73.56 & 10.49 & 69.84 & 24.09 & 52.64 \\
 & WINA          & 9.05 & 73.59 & 10.52 & 69.79 & 27.78 & 55.34 \\
 & TopK-Guided   & \textbf{9.04} & \textbf{73.88} & \textbf{10.46} & \textbf{70.02} & \textbf{21.55} & \textbf{57.85} \\
\bottomrule
\end{tabular}
\end{table}
\textbf{Setup:} We evaluate on Llama-2-7B and Llama-3-8B at target sparsity $s \in \{0.3, 0.5, 0.7\}$ on single NVIDIA A100 SXM4 — 40 GB GPU. We report C4 \citep{raffel2023exploringlimitstransferlearning} perplexity on 500 validation samples and average zero-shot accuracy over eight tasks (ARC-Easy \citep{clark2018thinksolvedquestionanswering}, WinoGrande \citep{sakaguchi2019winograndeadversarialwinogradschema}{}, HellaSwag \citep{zellers2019hellaswagmachinereallyfinish}, BoolQ \citep{clark2019boolqexploringsurprisingdifficulty}, SciQ \citep{welbl2017crowdsourcingmultiplechoicescience}, MMLU \citep{hendrycks2021measuringmassivemultitasklanguage}, GSM8K \citep{cobbe2021trainingverifierssolvemath}, LAMBADA \citep{paperno2016lambadadatasetwordprediction}) with the \texttt{lm-evaluation-harness} \citep{eval-harness}. Baselines follow their published setups: WINA uses tensor rotation with TopK masking, whereas TEAL uses threshold masking with no rotation. The two free hyperparameters $\bar{s}$ and $\lambda$ are selected on a held-out dataset, never on the reported one: for each target sparsity we sweep the grid on one dataset, keep the lowest-perplexity setting, and evaluate it on the other, so the reported C4 numbers use $(\bar{s}, \lambda)$ chosen on WikiText-2 (Appendix~\ref{app:param-sweep}).

\textbf{Main results:} Table~\ref{tab:main} reports C4 perplexity and average accuracy on both models. TopK-Guided attains the lowest perplexity and the highest accuracy in every one of the twelve comparisons. The margin over the baselines is small at $s=0.3$ and grows with sparsity, becoming largest at $s=0.7$: when most activations are kept, the masking rule and budget split have little room to act, but as sparsity rises each decision about \emph{which} entries to drop and \emph{where} to spend the budget matters more, and the two contributions have their greatest effect. One case stands out: on Llama-3-8B at $s=0.7$, TEAL records a lower perplexity than WINA yet a lower average accuracy, a reminder that perplexity and downstream accuracy need not move together. We do not investigate the cause of this reversal, but we note that it is exactly the kind of case where WINA is not uniformly the stronger baseline, and that TopK-Guided still surpasses both TEAL and WINA on both metrics there. An ablation (Appendix~\ref{app:ablation}) shows that both contributions help on their own and combine well, with the largest joint gain at $s=0.7$.

\textbf{Computational savings and quantization:} TopK-Guided delivers its gains at no added cost over WINA. Because it holds the realised per-token sparsity within $\pm\delta$ of the target, it skips the same fraction $s$ of each projection's computation as exact TopK, up to the small $\delta$ margin, for the same FLOP reduction; the only overhead is an $O(d)$ mask per token, negligible beside the projection. So at matched compute it improves on WINA on every setting, while also improving on threshold masking, whose realised sparsity can fall short of the target and deliver less than the promised saving, where our bounded deviation guarantees the target reduction is met. And it comes for free architecturally too: TopK-Guided operates on the same SVD-rotated weights WINA inherits from SliceGPT~\cite{ashkboos2024slicegptcompresslargelanguage}, the basis in which rotation-based quantization such as QuaRot~\cite{ashkboos2024quarotoutlierfree4bitinference} also acts, so the two compose on the same computationally-invariant model with no extra machinery.
\section{Conclusion and Limitations }
\label{sec:conclusion}

We presented TopK-Guided, two training-free changes to WINA that add no cost over it yet consistently improve on both WINA and TEAL. TopK-Guided masking recovers per-token adaptivity while keeping the realised sparsity within a bounded $\pm\delta$ of the target, and sensitivity-aware block sparsity splits a fixed global budget across blocks by a single-pass sensitivity estimate. The gains hold across both models and all sparsities and are largest where sparsification is most aggressive, and an ablation confirms that each contribution helps on its own. Our study has two main limits. We evaluate decoder-only models up to 8B parameters, so the method's behaviour at larger scale and on other architectures remains to be checked, though it makes no architecture-specific assumptions. We also report FLOP-level savings rather than wall-clock speedup, leaving a kernel-level implementation to future work. Finally, because TopK-Guided operates on the same rotated, computationally-invariant model as recent rotation-based quantization, combining the two is a natural next step toward faster low-precision inference.


\newpage
\bibliographystyle{plainnat}   
\bibliography{references}       

\newpage
\appendix
\section{Notation}
\label{app:notation}

\begin{table}[t]
\centering
\caption{The sparsity-related symbols used throughout the paper. }
\label{tab:notation}
\begin{tabular}{@{}l l p{0.62\linewidth}@{}}
\toprule
Symbol & Scope & Meaning \\
\midrule
$s$ & global & Target sparsity: the fraction we aim to zero, fixed for a run. \\
$\bar{s}$ & per-block & Probe sparsity used only to \emph{measure} a block's sensitivity in Eq.~\ref{eq:sensitivity}; never the level a block is finally run at (Sec.~\ref{sec:block}). \\
$e_i$ & per-block & Sensitivity of block $i$: its relative output error when sparsified at $\bar{s}$ (Eq.~\ref{eq:sensitivity}). \\
$s_i$ & per-block & Budget actually applied to block $i$, from redistributing $s$ across blocks by sensitivity (Eq.~\ref{eq:redistribution}). \\
$\hat{s}^{(t)}$ & per-token & Token $t$'s \emph{natural} sparsity: the fraction of its entries falling below $\tau$ (Eq.~\ref{eq:token-estimate}). \\
$s^{*(t)}$ & per-token & Target actually enforced for token $t$: $\hat{s}^{(t)}$ clipped to a $\pm\delta$ band around $s$ (Eq.~\ref{eq:clip}). \\
$\delta$ & global & Clip margin bounding $|s^{*(t)} - s| \le \delta$; fixed at $0.01$ (Eq.~\ref{eq:clip}). \\
$\lambda$ & global & Redistribution strength: how far $s_i$ moves from $s$ (Eq.~\ref{eq:redistribution}). \\
\bottomrule
\end{tabular}
\end{table}

Our method introduces several sparsity quantities that operate at different levels: a single global target, a per-block budget, and a per-token target, plus the calibration and control parameters that relate them. Because they share the symbol $s$, Table~\ref{tab:notation} collects them in one place, giving each symbol's scope, meaning, and the equation or section where it is defined.

\section{Target Sparsity Vs Realised Sparsity}
\label{app:threshold-vs-topk}

Threshold masking calibrates a fixed threshold $\tau$ offline to hit a target sparsity $s$, but because $\tau$ is frozen after calibration while each token's activation distribution differs, the fraction of activations actually falling below $\tau$ at inference varies from token to token and need not average to $s$. TopK masking, by retaining exactly $k = \lfloor d(1-s) \rfloor$ elements per token, hits the target by construction.

Table~\ref{tab:realised-sparsity} quantifies this on Llama-2-7B at a target of $s = 0.5$ across three datasets. TEAL's threshold masking consistently under-shoots the target, and by an amount that drifts from one dataset to the next, so its realised sparsity, and hence its actual speedup, is neither on target nor predictable. WINA's TopK hits the target exactly on every dataset, by construction. TopK-Guided stays within its $\pm\delta$ band of the target ($\delta = 0.01$) on all three datasets: it recovers threshold masking's per-token adaptivity while keeping the realised sparsity controlled, close to the target and stable across datasets, which is exactly what neither baseline achieves at once.

\begin{table}[h]
\centering
\caption{Target vs.\ realised sparsity of TEAL, WINA, and TopK-Guided on Llama-2-7B at target $s = 0.5$, over 500 samples from each dataset. Perplexity (PPL) is shown for reference.}
\label{tab:realised-sparsity}
\small
\begin{tabular}{l l c c c}
\toprule
Method & Dataset & Target $s$ & Realised & PPL \\
\midrule
TEAL & WikiText-2 & 50\% & 48.14\% & 7.45 \\
TEAL  & C4         & 50\% & 48.36\% & 7.57 \\
TEAL  & cc100 & 50\% & 47.64\% & 10.40 \\
\midrule
WINA      & WikiText-2 & 50\% & 50.00\% & 7.24 \\
WINA     & C4         & 50\% & 50.00\% & 7.34 \\
WINA       & cc100         & 50\% & 50.00\% & 10.28 \\
\midrule
TopK-Guided      & WikiText-2 & 50\% & 49.39\% & 7.20 \\
TopK-Guided      & C4         & 50\% & 49.52\% & 7.32 \\
TopK-Guided      & cc100         & 50\% & 49.46\% & 10.24 \\
\bottomrule
\end{tabular}
\end{table}
\section{Tensor Rotation: What, Why, and How}
\label{app:rotation}

Both WINA and our TopK-Guided method operate on a rotated version of the
model's weights and activations. This appendix explains what the rotation
is, why it makes WINA's weight-informed score well-founded, and how it is
applied at zero inference cost.

\subsection{The Problem: The Per-Element Bound Ignores Cross-Terms}

WINA's selection rule comes from a single-element error analysis: dropping
index $j$ perturbs the projection output by $h_j W_{:,j}$, with norm
$|h_j| \cdot \|W_{:,j}\|_2$. This per-element quantity is the score WINA
uses to rank activations.

At target sparsity $s$, however, many elements are dropped at once. Let
$\mathcal{Z}$ be the set of dropped indices, so $\hat{h}_j = 0$ for
$j \in \mathcal{Z}$ and $\hat{h}_j = h_j$ otherwise. The resulting error is
\begin{equation}
    e = Wh - W\hat{h} = \sum_{j \in \mathcal{Z}} h_j W_{:,j},
\end{equation}
the sum of each dropped element's individual contribution. Its squared
norm expands into a double sum over pairs of dropped indices, which splits
into diagonal and off-diagonal terms:
\begin{equation}
    \|Wh - W\hat{h}\|_2^2
    = \sum_{j \in \mathcal{Z}} h_j^2 \|W_{:,j}\|_2^2
    + \underbrace{\sum_{\substack{j \neq k \\ j,k \in \mathcal{Z}}}
      h_j h_k \langle W_{:,j}, W_{:,k} \rangle}_{\text{cross-terms}}.
    \label{eq:cross-terms}
\end{equation}
The first term is exactly the sum of WINA's squared per-element scores,
which is what its ranking implicitly optimises. The second term, the
cross-terms, depends on the inner products $\langle W_{:,j}, W_{:,k}
\rangle$ between the columns of dropped indices: if two dropped columns
are aligned and their activations share a sign, their errors reinforce and
the true error exceeds the per-element estimate. By summing only the
diagonal term, WINA implicitly assumes the columns of $W$ are orthogonal.
Learned weight matrices are not, so the cross-terms are non-negligible and
the per-element score does not reliably identify the elements whose joint
removal minimises output error.

\subsection{The Idea: Column-Orthogonalise the Weight Matrix}

If the columns of $W$ were orthogonal, then $\langle W_{:,j}, W_{:,k}
\rangle = 0$ for $j \neq k$, the cross-terms in Equation~\ref{eq:cross-terms}
would vanish, and the per-element score would sum to the true error.

WINA achieves this, following SliceGPT~\citep{ashkboos2024slicegptcompresslargelanguage}, by
right-multiplying each weight matrix with an orthogonal matrix $Q$ obtained
from its own Singular Value Decomposition. Writing $W = U\Sigma V^\top$
with $U, V$ orthogonal and $\Sigma$ diagonal, and choosing $Q = V$, the
transformed weight is
\begin{equation}
    \hat{W} = WQ = U\Sigma V^\top V = U\Sigma,
\end{equation}
whose columns are exactly orthogonal, since
$\hat{W}^\top \hat{W} = \Sigma^\top U^\top U \Sigma = \Sigma^2$ is
diagonal. Every off-diagonal inner product is zero by construction, so
WINA's per-element score applied to $\hat{W}$ recovers the true error
exactly.

This rotation is data-dependent, derived from an actual weight matrix
rather than fixed in advance. The SVD
construction yields exact orthogonality for the matrix whose SVD is taken,
at the cost of requiring the propagation machinery below to preserve the
model's output.

\subsection{The Implementation: Computational Invariance}

Orthogonalising a standalone layer $y = Wh$ via $\hat{W} = WQ$ would change
its output unless the input is also transformed to $Q^\top h$. In a
transformer, that input transformation must be produced by the preceding
layer, which then requires its own compensation, and so on. SliceGPT's key
observation is that this chain closes: the required rotations can be
absorbed into adjacent weight matrices throughout the network, so no
rotation is ever computed at runtime. RMSNorm is rotation-equivariant and
residual connections are linear, so both commute cleanly with $Q$.

After this one-time offline preprocessing, the rotated model produces
exactly the same outputs as the original (the computational invariance
property of SliceGPT) while performing the same number of FLOPs at
inference. All experiments in this paper, including sensitivity estimation and TopK-Guided masking, operate on these rotated weights.

\section{Block-wise Greedy Sparsity Allocation}
\label{app:greedy}

Both TEAL and WINA, and our method, allocate a block's sparsity budget
non-uniformly across its weight matrices using a greedy search. This
appendix describes that procedure, which our sensitivity-aware budgets
feed into unchanged.

\paragraph{Intuition:} Within a transformer block there are $n$ weight
matrices ($W_q, W_k, W_v, W_o$ in attention and $W_{\text{gate}},
W_{\text{up}}, W_{\text{down}}$ in the MLP). They are not equally
forgiving under sparsification: zeroing activations feeding into some
matrices barely changes the block's output, while the same fraction into
others causes large error. The greedy algorithm exploits this asymmetry.
It repeatedly asks which matrix can absorb a little more sparsity for the
smallest increase in output error, and assigns that matrix the next
increment. Repeating this traces out a good allocation for every overall
budget, computed once at calibration and looked up at inference.

\paragraph{Setup:} Let $f_i = \|W_i\|_0$ be the parameter count of matrix
$i$ and $F = \sum_i f_i$ the block total. Each matrix has its own sparsity
$p_i \in [0,1]$, and the block's compute-weighted sparsity is
$P = \sum_i p_i f_i / F$. Because larger matrices move $P$ more per unit of
$p_i$, each matrix takes a step size inversely proportional to its size,
\begin{equation}
    \delta_i = \alpha \cdot \frac{F}{f_i},
\end{equation}
where $\alpha$ is a small base step ($\alpha = 0.025$ in our experiments).
A small matrix thus takes large steps and a large matrix small steps, so a
single step from either moves the overall $P$ by roughly the same amount.

\paragraph{The loop:} Starting from $p_i = 0$ for all $i$, the algorithm
repeats until $P$ reaches $1$: (i) for every matrix $i$, tentatively raise
$p_i$ by $\delta_i$ and measure the output reconstruction error
$E_i = \|Y_{\text{gt}} - \hat{Y}_i\|_2$ against the dense block output
$Y_{\text{gt}}$, then undo the change; (ii) permanently commit the
increment for the matrix $j = \arg\min_i E_i$ with the least error; (iii)
recompute $P$ and record the row $(P, p_1, \dots, p_n)$ in the lookup
table. The full procedure is given in Algorithm~\ref{alg:greedy}.

\paragraph{Worked example:} Consider a toy block with two matrices: a
small matrix $A$ with $f_A = 100$ and a large matrix $B$ with $f_B = 400$
(so $F = 500$), using $\alpha = 0.1$. The step sizes are
$\delta_A = 0.1 \cdot 500/100 = 0.5$ and
$\delta_B = 0.1 \cdot 500/400 = 0.125$, so $A$ moves in large jumps and
$B$ in small ones. Suppose the errors evolve as in
Table~\ref{tab:greedy-toy}. At steps 1 and 2, sparsifying $B$ by its small
step costs far less than sparsifying $A$ by its large one, so $B$ is
chosen. By step 3, $B$ has absorbed two increments and its next one costs
more than giving $A$ its first, so the algorithm switches to $A$. After
three steps the block reaches $P = 0.3$ with the non-uniform allocation
$p_A = 0.5, p_B = 0.25$, an allocation a uniform ``sparsify everything by
30\%'' baseline would not find. Over the real $n = 7$ matrices, the large
MLP matrices tend to absorb sparsity earliest, while smaller, more
sensitive attention projections are protected until later.

\begin{table}[h]
\centering
\caption{Toy trajectory of Algorithm~\ref{alg:greedy} on a 2-matrix block
($f_A{=}100$, $f_B{=}400$, $\alpha{=}0.1$).}
\label{tab:greedy-toy}
\small
\begin{tabular}{c c c c c c}
\toprule
Step & Candidate errors $(E_A, E_B)$ & Winner & $p_A$ & $p_B$ & $P$ \\
\midrule
0 & --            & --  & 0.000 & 0.000 & 0.000 \\
1 & $(2.0,\,0.5)$  & $B$ & 0.000 & 0.125 & 0.100 \\
2 & $(2.0,\,1.3)$  & $B$ & 0.000 & 0.250 & 0.200 \\
3 & $(2.0,\,2.5)$  & $A$ & 0.500 & 0.250 & 0.300 \\
\bottomrule
\end{tabular}
\end{table}

\begin{algorithm}[H]
\caption{Block-wise Greedy Sparsity Allocation}
\label{alg:greedy}
\begin{algorithmic}[1]
\Require Block $\mathcal{B}$, base step $\alpha$, calibration input $X \in \mathbb{R}^{B \times \mathrm{seq} \times d}$, weight matrices $W_1,\ldots,W_n$
\State $f_i \leftarrow \|W_i\|_0$ \quad for $i = 1,\ldots,n$ \Comment{parameter count of each matrix}
\State $F \leftarrow \sum_{i=1}^n f_i$ \Comment{total parameter count of block}
\State $\mathbf{p} \leftarrow \mathbf{0}_n$,\quad $P \leftarrow 0$ \Comment{initialise all sparsities to zero}
\State $Y_{\mathrm{gt}} \leftarrow \mathcal{B}(X)$ \Comment{dense forward pass — ground truth output}
\While{$P < 1$}
    \For{$i = 1$ \textbf{to} $n$}
        \State $\delta_i \leftarrow \alpha \cdot (F / f_i)$
        \State $p_i \leftarrow p_i + \delta_i$ \Comment{tentatively increase sparsity of $W_i$}
        \State $\hat{Y}_i \leftarrow \mathcal{B}(X,\, \mathbf{p})$
        \State $E_i \leftarrow \|\,Y_{\mathrm{gt}} - \hat{Y}_i\,\|_2$
        \State $p_i \leftarrow p_i - \delta_i$ \Comment{undo tentative increment}
    \EndFor
    \State $j \leftarrow \arg\min_i\, E_i$ \Comment{projection with least reconstruction error}
    \State $p_j \leftarrow p_j + \delta_j$ \Comment{permanently commit the increment}
    \State $P \leftarrow \displaystyle\sum_{i=1}^n (p_i \cdot f_i)\,/\,F$
    \State \textbf{record} $(\mathbf{p},\, P)$ \Comment{append row to lookup table}
\EndWhile
\end{algorithmic}
\end{algorithm}

\section{Hyperparameter Sweep }
\label{app:param-sweep}

With the clip margin fixed at $\delta = 0.01$, TopK-Guided has two free
hyperparameters: the probe sparsity $\bar{s}$ used for sensitivity
estimation and the redistribution strength $\lambda$. This appendix
reports WikiText-2 perplexity for every combination in the grid
$\bar{s}, \lambda \in \{0.2, 0.4, 0.6, 0.8\}$ at each target sparsity
$s \in \{0.3, 0.5, 0.7\}$ on Llama-2-7B. In each table, entries
that meet or beat the WINA baseline for that target sparsity are shown in
bold. The purpose is to establish whether a consistent parameter region
outperforms the baselines, so that the method can be applied without a
careful per-setting search.

\begin{table}[h]
\centering
\caption{WikiText-2 perplexity on Llama-2-7B across the $(\bar{s}, \lambda)$ grid
for each target sparsity. Bold entries meet or beat the WINA baseline for
that sparsity. Columns are $\lambda \in \{0.2, 0.4, 0.6, 0.8\}$; rows are
$\bar{s} \in \{0.2, 0.4, 0.6, 0.8\}$.}
\label{tab:param-sweep}
\small
\begin{tabular}{c cccc}
\toprule
$\bar{s} \backslash \lambda$ & 0.2 & 0.4 & 0.6 & 0.8 \\
\midrule
\multicolumn{5}{c}{\textit{$s=0.3$ \quad (TEAL $=6.83$, WINA $=6.78$)}} \\
\midrule
0.2 & \textbf{6.77} & \textbf{6.77} & \textbf{6.77} & \textbf{6.77} \\
0.4 & \textbf{6.77} & \textbf{6.78} & \textbf{6.77} & \textbf{6.78} \\
0.6 & \textbf{6.78} & \textbf{6.78} & \textbf{6.78} & \textbf{6.78} \\
0.8 & \textbf{6.78} & \textbf{6.78} & \textbf{6.78} & 6.80 \\
\midrule
\multicolumn{5}{c}{\textit{$s=0.5$ \quad (TEAL $=7.45$, WINA $=7.24$)}} \\
\midrule
0.2 & \textbf{7.22} & \textbf{7.20} & \textbf{7.22} & \textbf{7.21} \\
0.4 & \textbf{7.21} & \textbf{7.22} & \textbf{7.23} & \textbf{7.23} \\
0.6 & \textbf{7.22} & \textbf{7.23} & 7.25 & 7.27 \\
0.8 & \textbf{7.21} & 7.26 & 7.28 & 7.32 \\
\midrule
\multicolumn{5}{c}{\textit{$s=0.7$ \quad (TEAL $=17.90$, WINA $=10.79$)}} \\
\midrule
0.2 & \textbf{10.49} & \textbf{10.29} & \textbf{10.16} & \textbf{10.14} \\
0.4 & \textbf{10.24} & \textbf{10.13} & \textbf{10.09} & \textbf{10.09} \\
0.6 & \textbf{10.23} & \textbf{10.07} & \textbf{10.11} & \textbf{10.34} \\
0.8 & \textbf{10.16} & \textbf{10.14} & \textbf{10.37} & 10.87 \\
\bottomrule
\end{tabular}
\end{table}

\paragraph{Observations:} At $s{=}0.3$, all 16 combinations beat TEAL and
15 of 16 meet or beat WINA; perplexity varies over just 0.03 points
(6.77 to 6.80), so at low sparsity the method is essentially insensitive
to the parameter choice. At $s{=}0.5$, all 16 beat TEAL, and 11 of 16 beat WINA. At $s{=}0.7$, all 16 combinations beat TEAL,
15 of 16 beat WINA. Across all three sparsity levels
the region $\bar{s} \in [0.2, 0.6]$, $\lambda \in [0.2, 0.4]$ is
consistently strong, indicating that TopK-Guided does not require an
extensive hyperparameter search in practice.

\section{Ablation of our Contributions}
\label{app:ablation}

To verify that each contribution helps on its own, we evaluate four
configurations on Llama-2-7B, starting from the WINA baseline and adding
each component in turn (Table~\ref{tab:ablation}). The first row is WINA.
The second adds only sensitivity-aware block sparsity, keeping WINA's
TopK masking. The third adds only TopK-Guided masking, keeping WINA's
uniform block budget. The fourth is the full method, with both
contributions active.

As in the main results, the two hyperparameters $(\bar{s}, \lambda)$ are
tuned on one dataset and the resulting fixed setting is then evaluated on
the other. Concretely, for each target sparsity we sweep the
$(\bar{s}, \lambda)$ grid on the selection dataset and keep the single
combination that gives the lowest perplexity there. That one chosen
setting is then held fixed and used to produce the reported numbers on the
other dataset, which took no part in the selection. We show both directions: the C4 columns
use the $(\bar{s}, \lambda)$ selected on WikiText-2, and the WikiText-2
columns use the $(\bar{s}, \lambda)$ selected on C4. The selection is
repeated independently for each target sparsity.

\begin{table}[h]
\centering
\caption{Ablation on Llama-2-7B. Each row adds one contribution on top of
WINA; perplexity is lower-is-better. Hyperparameters are selected on the
opposite dataset to the one reported (Appendix~\ref{app:param-sweep}).
Best per column in bold.}
\label{tab:ablation}
\small
\begin{tabular}{l ccc ccc}
\toprule
& \multicolumn{3}{c}{C4 (selected on WikiText-2)}
& \multicolumn{3}{c}{WikiText-2 (selected on C4)} \\
\cmidrule(lr){2-4}\cmidrule(lr){5-7}
Configuration & $s{=}0.3$ & $s{=}0.5$ & $s{=}0.7$
              & $s{=}0.3$ & $s{=}0.5$ & $s{=}0.7$ \\
\midrule
WINA (baseline)
 & 6.90 & 7.34 & 10.72 & 6.78 & 7.24 & 10.79 \\
\; + Sensitivity-aware sparsity
 & 6.90 & 7.35 & 10.25 & 6.78 & 7.24 & 10.23 \\
\; + TopK-Guided masking
 & \textbf{6.89} & 7.33 & 10.66 & \textbf{6.77} & 7.21 & 10.67 \\
\; + Both (TopK-Guided)
 & \textbf{6.89} & \textbf{7.32} & \textbf{10.15}
 & \textbf{6.77} & \textbf{7.20} & \textbf{10.09} \\
\bottomrule
\end{tabular}
\end{table}

\textbf{Observations:} Each contribution independently matches or improves on WINA in most settings. Sensitivity-aware block sparsity tracks WINA closely at low sparsity and gives its clearest gain at $s=0.7$, where budget redistribution matters most. TopK-Guided masking alone meets or beats WINA across the board, with modest gains that grow with sparsity. Combined, the two reinforce each other: the full method is best at every sparsity level, with the largest joint gain at $s=0.7$. The two effects are largely complementary, so the full method tracks the better of the two at each level.

\section{Per-Task Downstream Accuracy}
\label{app:downstream-full}

Tables~\ref{tab:downstream-llama2} and~\ref{tab:downstream-llama3} give
the full per-task breakdown of the downstream accuracy summarised in the
main text, for Llama-2-7B and Llama-3-8B respectively. All tasks are
evaluated zero-shot with the \texttt{lm-evaluation-harness}. The average
column matches the one reported in the main results.

\begin{table}[h]
\centering
\caption{Per-task accuracy (\%) on Llama-2-7B across eight tasks. Best
average per sparsity level in bold.}
\label{tab:downstream-llama2}
\small
\setlength{\tabcolsep}{4pt}
\begin{tabular}{l cccccccc c}
\toprule
Method & ARC-E & WG & HS & BoolQ & SciQ & MMLU & GSM8K & LAMBADA & Avg \\
\midrule
\multicolumn{10}{l}{\textit{Sparsity $s=0.3$}} \\
TEAL         & 74.24 & 68.98 & 76.02 & 77.19 & 91.20 & 41.12 & 13.42 & 73.51 & 64.46 \\
WINA         & 74.49 & 68.51 & 76.01 & 77.06 & 91.00 & 42.11 & 14.33 & 73.55 & 64.63 \\
TopK-Guided  & 74.71 & 68.90 & 75.89 & 77.22 & 91.10 & 41.97 & 14.33 & 73.57 & \textbf{64.71} \\
\midrule
\multicolumn{10}{l}{\textit{Sparsity $s=0.5$}} \\
TEAL         & 72.64 & 66.30 & 73.87 & 75.60 & 90.40 & 36.48 & 10.24 & 69.49 & 61.88 \\
WINA         & 73.44 & 68.51 & 74.42 & 75.84 & 91.00 & 38.98 & 11.22 & 73.22 & 63.33 \\
TopK-Guided  & 73.11 & 68.51 & 74.43 & 75.44 & 91.50 & 39.29 & 11.98 & 73.14 & \textbf{63.42} \\
\midrule
\multicolumn{10}{l}{\textit{Sparsity $s=0.7$}} \\
TEAL         & 62.33 & 59.75 & 60.61 & 65.99 & 81.60 & 26.34 & 1.67 & 43.22 & 50.19 \\
WINA         & 66.04 & 64.09 & 64.90 & 70.49 & 89.20 & 28.86 & 2.65 & 66.10 & 56.54 \\
TopK-Guided  & 65.32 & 64.09 & 66.33 & 70.24 & 88.10 & 29.45 & 3.18 & 67.65 & \textbf{56.80} \\
\bottomrule
\end{tabular}
\end{table}

\begin{table}[h]
\centering
\caption{Per-task accuracy (\%) on Llama-3-8B across eight tasks. Best
average per sparsity level in bold.}
\label{tab:downstream-llama3}
\small
\setlength{\tabcolsep}{4pt}
\begin{tabular}{l cccccccc c}
\toprule
Method & ARC-E & WG & HS & BoolQ & SciQ & MMLU & GSM8K & LAMBADA & Avg \\
\midrule
\multicolumn{10}{l}{\textit{Sparsity $s=0.3$}} \\
TEAL         & 77.06 & 72.38 & 78.72 & 81.07 & 93.90 & 61.32 & 47.92 & 76.11 & 73.56 \\
WINA         & 77.31 & 73.16 & 78.80 & 81.41 & 93.70 & 61.47 & 47.01 & 75.86 & 73.59 \\
TopK-Guided  & 77.23 & 73.48 & 78.89 & 81.59 & 93.80 & 61.61 & 48.29 & 76.13 & \textbf{73.88} \\
\midrule
\multicolumn{10}{l}{\textit{Sparsity $s=0.5$}} \\
TEAL         & 76.14 & 70.40 & 75.88 & 79.45 & 93.30 & 57.16 & 32.75 & 73.61 & 69.84 \\
WINA         & 75.42 & 71.74 & 76.18 & 81.96 & 92.80 & 56.78 & 29.04 & 74.40 & 69.79 \\
TopK-Guided  & 75.76 & 70.24 & 76.24 & 81.50 & 93.60 & 57.42 & 30.63 & 74.77 & \textbf{70.02} \\
\midrule
\multicolumn{10}{l}{\textit{Sparsity $s=0.7$}} \\
TEAL         & 62.21 & 61.72 & 59.77 & 67.13 & 86.60 & 31.75 & 3.56 & 48.38 & 52.64 \\
WINA         & 62.08 & 60.77 & 60.62 & 72.57 & 88.80 & 35.29 & 3.03 & 59.58 & 55.34 \\
TopK-Guided  & 64.69 & 62.27 & 63.72 & 73.18 & 89.80 & 42.91 & 4.09 & 62.14 & \textbf{57.85} \\
\bottomrule
\end{tabular}
\end{table}
\FloatBarrier
\section{Impact Statement}
\label{app:impact}
This work aims to reduce the computational cost of LLM inference, which may lower energy use and make LLM deployment more accessible in resource-constrained settings. We do not identify any direct negative societal impacts specific to the proposed method, as it does not introduce new model capabilities, training data, or user-facing behavior.


\end{document}